\documentclass{ifacconf}

\usepackage{graphicx}      %
\usepackage{natbib}		   %
\usepackage{amsmath,amssymb,amsfonts}      %
\usepackage{xcolor}

\newcommand{\abs}[1]{\left\lvert#1\right\rvert}

\newcommand{\defeq}{\doteq}
\newcommand{\confreg}[1][2]{\mathcal{C}_{#1}}
\newcommand{\outappr}[1][2]{\mathcal{O}_{#1}}

\newlength{\dhatheight}

\begin{document}
	\begin{frontmatter}

		\title{Sample Complexity of the Sign-Perturbed Sums Identification Method: Scalar Case\hspace{-0.3mm}\thanksref{footnoteinfo}} 

		\thanks[footnoteinfo]{This research was supported by the European Union within the framework of the National Laboratory for Autonomous Systems, RRF-2.3.1-21-2022-00002; and 
		by the TKP2021-NKTA-01 grant of the National Research, Development and Innovation Office, Hungary.
		}
		
		\author[First]{Szabolcs Szentpéteri}\qquad 
		\author[First,Second]{Balázs Csan\'ad Csáji} 
		
		\address[First]{Institute for Computer Science and Control (SZTAKI),\\
		Eötvös Loránd Research Network (ELKH),\\
		13-17 Kende utca, H-1111, Budapest, Hungary}
		\address[Second]{Institute of Mathematics, E\"otv\"os Lor\'and University (ELTE),\\
		1/C Pázmány Péter sétány, H-1117, Budapest, Hungary \\[1mm]
		emails: \{szentpeteri.szabolcs, csaji.balazs\}@sztaki.hu
		}
		
		\begin{abstract}                %
			Sign-Perturbed Sum (SPS) is a powerful finite-sample system identification algorithm which can construct confidence regions for the true data generating system with exact coverage probabilities, for any finite sample size. SPS was developed in a series of papers and it has a wide range of applications, from general linear systems, even in a closed-loop setup, to nonlinear and nonparametric approaches. Although several theoretical properties of SPS were proven in the literature,
			the sample complexity of the method was not analysed so far. This paper aims to fill this gap and provides the first results on the sample complexity of SPS. Here, we focus on scalar linear regression problems, that is we study the behaviour of SPS confidence intervals.
			We provide high probability upper bounds, under three different sets of assumptions, showing that the 
			sizes of SPS confidence intervals shrink at a geometric rate around the true parameter, 
			if the observation noises are subgaussian.
			We also show that similar bounds hold for the previously proposed outer approximation of the confidence region.
			Finally, we present simulation experiments comparing the theoretical and the empirical convergence rates.
		\end{abstract}
		
		\begin{keyword}
			Randomized methods for modeling, identification and signal processing
		\end{keyword}
		
	\end{frontmatter}

	\section{Introduction}
	\vspace*{-2mm}
	Estimating models is a fundamental problem across several domains, such as system identification, machine learning and statistics. System identification, a research area that studies how to build 
	models of dynamical systems from observed data, has a long and rich history. Classical methods in the field provide asymptotically guaranteed estimates and confidence regions \citep{Soderstrom1989,Ljung1999}. Recently, a paradigm shift took place in the field
	and more significant emphasis was given to approaches with non-asymptotic guarantees. Most of these techniques assume that the noises and disturbances follow given (known) distributions, therefore distribution-free, non-asymptotic identification of dynamical systems still remain an active area of research \citep{Algo2018}. 
	
	Two promising identification algorithms that can construct non-asymptotic confidence regions around the true parameter, for any finite sample, in a distribution-free setting are the LSCR: Leave-out Sign-dominant Correlation Regions \citep{Campi2005} and the SPS: Sign-Perturbed Sums \citep{Csaji2015} methods.
	
	SPS constructs exact confidence regions around the least-squares estimate for any finite sample, under mild assumptions on the noises, namely that they are independent and their probability distributions are symmetric about zero.
	
	Several important properties of SPS, such as its exact coverage probability \citep{Csaji2015} and strong consistency \citep{Weyer2017}, were rigorously proven for linear regression problems, under the assumptions mentioned above. The standard SPS method provides an indicator function, which evaluates whether a given parameter is included in the confidence region. In \citep{Csaji2015} an ellipsoidal outer approximation algorithm was proposed that builds a compact representation of the confidence set around the least-squares estimate. The symmetricity assumption on the noises was relaxed in \citep{kolumban2015perturbed}.
	The closed-loop applicability of SPS was studied in \citep{Csaji2015cdc}, and in \citep{volpe2015sign} an instrumental variable based generalization was given that can construct confidence regions for ARX systems with the same theoretical guarantees as mentioned before. The behaviour of SPS was also investigated in the face of undermodelling \citep{Care2021}. 
	Further extensions and applications of SPS include kernel-based methods \citep{csaji2019distribution,baggio2022bayesian}, nonparametric confidence bands \citep{csaji2022nonparametric} 
	and even tests for binary classification \citep{tamas2021exact}.
	
	Although the confidence regions generated by SPS are strongly consistent, the sample complexity of the method remained an open question. Here, we give  
	distribution-free, high probability bounds for the length of SPS confidence intervals for any finite sample size. As a first line of sample complexity research for SPS, the emphasis of our study is on the possible tools of theoretical analysis, hence, we restrict our attention to the scalar-valued case. Although we only investigate the scalar setting, the obtained results are directly relevant, for example, for 
    multi-armed bandit problems \citep{lattimore_szepesvari_2020}, 
    prediction intervals, confidence bands \citep{csaji2022nonparametric} 
    and signal processing approaches \citep{Csaji2011}. 
    
    Our main contributions in this paper are as follows:
	\begin{enumerate}
		\item Non-asymptotic analysis of SPS in case of the ``constant in noise'' setting, assuming subgaussian noises.
		\item High probability upper bounds for the sizes of SPS confidence intervals for scalar linear regression, both for deterministic and stochastic regressors.
		\item Simulation experiments to compare the obtained theoretical bounds with the empirical performance.
	\end{enumerate}
	The paper is structured as follows. In Section \ref{sec:SPS_overview} we give a short overview of SPS and its fundamental properties. Section \ref{sec:SPS_sample_complex} provides our theoretical non-asymptotic results on the size of the SPS confidence regions with proofs. The simulation experiments are presented in Section \ref{sec:simulation}. Finally, Section \ref{sec:conclusion} summarizes and concludes the paper.
	
	\section{The Sign-Perturbed Sums algorithm}\label{sec:SPS_overview}
	\vspace{-2mm}
	In this section we give an overview of SPS for scalar linear regression. The reader is referred to \citep{Csaji2015} and \citep{Weyer2017} for a detailed description of the algorithm in the general (d-dimensional) linear regression case, with theorems and proofs. Note that in our study we discard (w.l.o.g.) the ``shaping matrix'' term of the algorithm, since its purpose is take the inter-dependencies of the parameteres into account, in case $d > 1$. In the scalar case this does not affect the constructed intervals.
	
	\subsection{Problem setting and main assumptions}
	\vspace{-2mm}
	Consider the following scalar linear regression system
	\vspace{1mm}
	\begin{equation}
		y_t = \varphi_t\hspace{0.3mm}\theta^* + w_t,
        \vspace{1mm}		
	\end{equation}
	where $\varphi_t \in \mathbb{R}$ is the regressor, $y_t \in \mathbb{R}$ is the output, $w_t \in \mathbb{R}$ is the noise and $\theta^* \in \mathbb{R}$ is the (constant) ``true'' parameter to be estimated. We are given a sample of size $n$ which consists of $\varphi_1, \dots, \varphi_n$ (inputs) and $y_1, \dots, y_n$ (outputs).
	
	The assumptions on the noises and the regressors are
	\begin{itemize}
		\item[A1] {\em $\{w_t\}$ is a sequence of independent random variables and each $w_t$ has a symmetric probability distribution about zero {\em(}i.e., $w_t$ has the same distribution as $-w_t${\em)}.}\\[-2mm]
		\item[A2] {\em The regressors, $\{\varphi_t\}$, are almost surely nonzero random variables, and $\{\varphi_t\}$ is independent of $\{w_t\}$.}
	\end{itemize}
	
	\subsection{The SPS algorithm and its theoretical properties}
	\vspace{-2mm}
	In linear regression problems given a sample of size $n$ the least-squares estimate (LSE) can be obtained by solving the normal equation. The core idea behind SPS is to introduce $m-1$ sign-perturbed sums $\{S_i(\theta)\}$ and a reference sum $S_0(\theta)$ from the normal equation and construct a confidence region based on the rank of $S_0(\theta)$.
	
\begin{figure}[t]
			\includegraphics[width=8.6cm]{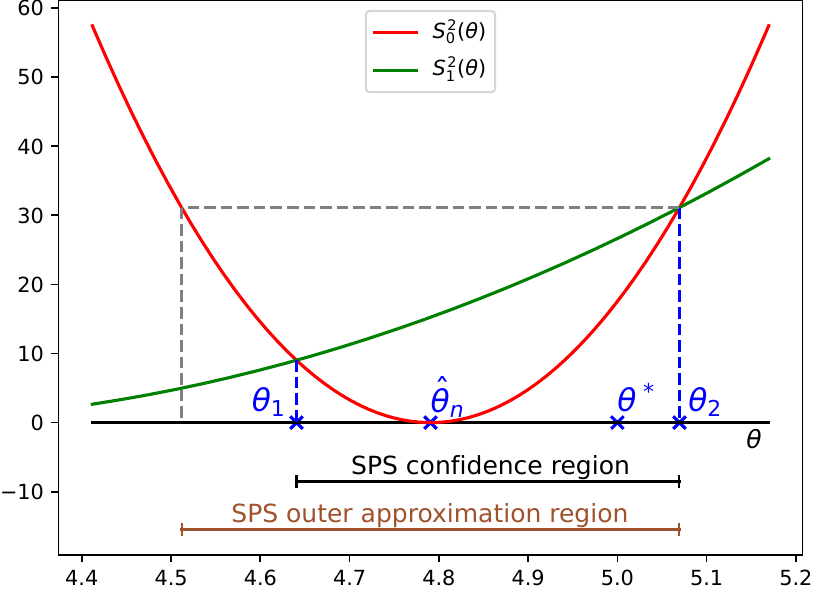}    %
			\caption{SPS confidence intervals, $m=2$ ($50\,\%$ confidence).} 
			\label{fig:SPS_explain}
	\end{figure}

	The SPS algorithm consists of two parts, an initialization phase and an indicator function. In the initialization part the algorithm calculates the main parameters and generates the random signs needed for the construction of the confidence region. The indicator function evaluates whether a given parameter $\theta$ is included in the confidence interval. The initialization is shown in Table \ref{spsinit} and the indicator is presented in Table \ref{spsindicator}.
	\begin{table}[t]
		\centering
		\caption{Pseudocode: SPS-Initialization $(\hspace{0.2mm}p\hspace{0.2mm})$}
		\vspace{-1.5mm}
		\label{spsinit}
		\setlength{\tabcolsep}{3pt}
		\begin{tabular}{p{10pt}p{205pt}}
			\hline
			1.& 
			Given a (rational) confidence probability $p \in (0,1)$, set integers $m > q >0$ such that $p = 1 - q/m$; \\
			2.& 
			Generate $n(m-1)$ i.i.d random signs $\{\alpha_{i,t}\}$ with
			\[\mathbb{P}(\alpha_{i,t} = 1) = \, \mathbb{P}(\alpha_{i,t} = -1) = 0.5\] 
			for all integers $1 \leq i \leq m-1$ and $1 \leq t \leq  n$.\\
			3.&
			Generate a permutation $\pi$ of the set $\{0,\dots, m - 1\}$ randomly, where each of the $m!$ possible permutations has the same probability $1/(m!)$ to be selected.\\
			\hline
		\end{tabular}
	\end{table}
	\begin{table}[t]
		\centering
		\caption{Pseudocode: SPS-Indicator $(\hspace{0.2mm}\theta\hspace{0.2mm})$}
		\vspace{-2.5mm}		
		\label{spsindicator}
		\setlength{\tabcolsep}{3pt}
		\begin{tabular}{p{10pt}p{205pt}}
			\hline
			1.& 
			For a given $\theta$, compute the prediction errors\vspace{-0.5mm}
			\[ \varepsilon_t(\theta) \defeq y_t - \varphi_t\theta\vspace{-1mm} \] 
			for all $1 \leq t \leq  n$;\\
			2.& 
			Evaluate\vspace{-1mm}
			\[ S_0(\theta) \defeq \sum_{t=1}^{n}\varphi_t\varepsilon_t(\theta), \quad \text{and} \quad
			S_i(\theta) \defeq \sum_{t=1}^{n}\alpha_{i,t}\varphi_t\varepsilon_t(\theta),\vspace{-0.5mm} \]
			for all indices $1 \leq i \leq m-1$; \\
			3.& 
			Order scalars $\{S_i^2(\theta)\}$ according to $\succ_{\pi}$, where ``$\succ_{\pi}$'' is ``$>$'' with random tie-breaking \citep{Csaji2015}; \\
			4.& 
			Compute the rank $\mathcal{R}(\theta)$ of $S_0^2(\theta)$ in the ordering where \vspace{-1.5mm}
			\begin{equation*}
				\mathcal{R}(\theta) \defeq \Bigg[1+\sum_{i=1}^{m-1}\mathbb{I}\left(S_0^2 (\theta)\succ_{\pi} S_i^2(\theta)\right)\Bigg];
				\vspace{-2mm}
			\end{equation*}\\
			5.&
			Return 1 if $\mathcal{R}(\theta) \leq m - q$, otherwise return 0.\\
			\hline
		\end{tabular}
	\end{table}
	Using this construction, the $p$-level SPS confidence region can be defined as
    \vspace{1mm}
	\begin{equation}
	    \confreg[p]\, \defeq\, \{\,\theta \in \mathbb{R}\text{ : SPS-Indicator}(\theta) = 1\,\}.
    \vspace{1mm}		
	\end{equation}
	As it was shown for general linear regression problems in \citep{Csaji2015}, the confidence region $\confreg[p]$ contains the true parameter $\theta^*$ exactly with probability $p$, thus 
	\begin{thm}\label{thm:exact_confidence}
		{\em Assuming A1 and A2, the coverage probability of the SPS confidence interval is exactly $p$, that is,}
		\begin{equation}
			\mathbb{P}(\theta^* \in \confreg[p])\, =\, 1-\frac{q}{m}\, =\, p.
		\vspace{-1mm}
		\end{equation}
	\end{thm}
	Note that in \citep{Csaji2015} 
	this theorem is proved for deterministic regressors, but it is straightforward to generalize the result to the case  of random regressors that are independent of the noises (i.e., by conditioning on the regressors, as the deterministic result can be applied to almost all realizations of the regressors). In \citep{Weyer2017} it has been rigorously proved that the confidence regions are also strongly consistent,
    which requires some further mild assumptions that we do not detail here.
	To give a compact representation of the confidence region around the LSE, an ellipsoidal outer approximation method was developed \citep{Csaji2015}. The confidence interval given by the outer approximation in our case is
	\begin{equation}
		\confreg[p]\, \subseteq\, \outappr[p]\, \defeq\, \big\{\,\theta \in \mathbb{R} : (\theta - \hat{\theta}_n)^2 r_n \leq r\,\big\},
		\vspace{1mm}	
	\end{equation}
	where $r_n = \tfrac{1}{n}\sum_{t=1}^{n}\varphi_t^2$ and $r$ can be calculated from the (ordering of the) solutions of the following optimization problems, for all $0 < i < m$, \citep{Csaji2015}
	\begin{equation}\label{equ:SPS_EOA}
		\begin{aligned}
			\text{maximize} \quad & S_0^2(\theta)\\[0.5mm]
			\textrm{subject to} \quad &S_0^2(\theta) - S_i^2(\theta) \leq 0.\\[1mm]
		\end{aligned}
	\end{equation}
	An illustrative example of the scalar SPS confidence regions for $m=2$ and $\forall\, t:\varphi_t =1$, is presented in Fig. ~\ref{fig:SPS_explain}. The confidence region consists of the points $\{\theta \in \mathbb{R}: S_0^2(\theta) \leq S_1^2(\theta)\}$. In the case of outer approximation, from \eqref{equ:SPS_EOA} we have $\sqrt{r} = \max(|\theta_{1} - \hat{\theta}_n|, |\theta_{2} - \hat{\theta}_n|)$, hence, the outer approximation is given by $\{\theta \in \mathbb{R}: |\theta - \hat{\theta}_n| \leq \sqrt{r}\}$.
	
	In case of a scalar parameter, the SPS confidence regions are in fact intervals. Thus, they automatically have nice representations, hence their outer approximations look superfluous (unlike in higher dimensions). Nevertheless, we also study the behaviour of the outer approximations, since we want to understand their sample complexity, as well, as they are very important for multi-dimensional settings.

	\section{Sample complexity of SPS}\label{sec:SPS_sample_complex}
    \vspace{-2mm}
	In this section we prove high probability upper bounds for the lengths of SPS confidence intervals. We analyze SPS assuming a one-dimensional constant parameter, first without inputs. Then, we investigate 
	linear regression with scalar inputs, using two other sets of assumptions.
	\subsection{Identifying a ``constant in noise''}
	\vspace{-2mm}
	Consider the problem of identifying a constant in noise
	\vspace{0.5mm}
	\begin{equation}\label{equ:param_idenfy_sys}
		y_t = \theta^* + w_t,
    \vspace{0.5mm}		
	\end{equation}
	for $t= 1, \dots, n$, where $y_t$ is the output, $w_t$ is the noise and $\theta^*$ is the true parameter (constant).
	Both the random variables $y_t$, $w_t$ and the optimal parameter $\theta^*$ are scalars.
	
	For this analysis, we also assume that
	\begin{itemize}
		\item[A3] {\em Every noise term, $w_t$, has a nonatomic, subgaussian distribution with variance proxy $\sigma^2$, that is $\forall\,\lambda \in \mathbb{R}:$
        \begin{align}
        \mathbb{E}\big[\exp(\lambda w_t)\hspace{0.3mm}\big]\, \leq\, \exp\!{\left(\frac{\lambda^2\sigma^2}{2}\right)}.
        \end{align}
        }
	\end{itemize} 
	\smallskip

	Note that this assumption is much weaker than assuming Gaussian noises. For example, every distribution with bounded support is automatically subgaussian, therefore, it covers a wide range of possible noises. The nonatomicity of the distributions is nonessential, it is only used to avoid ties and, thus, to simplify the analysis. Moreover, SPS does not exploit subgaussianity in any way, A3 is used solely for the sake of studying the sample complexity of SPS.

	\citet{Care2022} has shown that the SPS confidence regions are bounded, if both the perturbed and the unperturbed regressors span the whole space. In our current setting, this means that there should be at least one positive and one negative sign in each sign-perturbation sequence. Thus, in order to guarantee boundedness, we assume
	
	\begin{itemize}
		\item[A4] {\em For all\, $0<i<m$, we have $\alpha_{i,1} = 1$ and $\alpha_{i,2} = -1$.}
	\end{itemize}
	
	Before we state our theorem, we first prove a lemma that is an essential part of the proof of the later theorem.
    \begin{lem}\label{lemma:intersection_is_sg}
		{\em Assume A1 and A3, $n > 2$, and let\vspace{1mm}
		$$X_+ \defeq \frac{\sum_{t=1}^{n}(1+\alpha_t)w_t}{\sum_{t=1}^{n}(1+\alpha_t)}, \qquad X_- \defeq \frac{\sum_{t=1}^{n}(1-\alpha_t)w_t}{\sum_{t=1}^{n}(1-\alpha_t)},\vspace{1mm}$$
        where $\alpha_1 = 1$, $\alpha_2 = -1$, and $\{\alpha_t\}_{t=3}^n$ are i.i.d.\ Rademacher random variables, independent of the noise terms $\{w_t\}$.
        
        Then, variables $X_+$ and $X_-$ satisfy for all $\varepsilon \geq 0:$
        \begin{equation}
        \mathbb{P}\left( |X_{\pm}| \geq \varepsilon\right) \, \leq\, 2 \left(\frac{1}{2}+\frac{1}{2}\exp\left(-\frac{\varepsilon^2}{2\sigma^2} \right) \right)^{\!n-1}\hspace{-5mm}.
        \vspace{-2mm}
        \end{equation}}
	\end{lem}
	\begin{pf}
		Observe that $\sum_{t=1}^{n}(1-\alpha_t) \overset{d}{=} \sum_{t=1}^{n}(1+\alpha_t)$, therefore it is enough to prove the claim for $X_+$, which in this proof we will denote by $X$ for shorthand of notation.

        Let $Z = \sum_{t=3}^{n}(1+\alpha_t)/2$, which has a binomial distribution with parameters $(1/2,n-2)$. By using the law of total probability, 
        the following can be written
        \begin{align}\label{equ:lemma2_first_part}
            &\mathbb{P}\left( |X| \geq \varepsilon\right) \,=\, \mathbb{P}\left( \Bigg|\,\frac{\sum_{t=1}^{n}(1+\alpha_t)w_t}{\sum_{t=1}^{n}(1+\alpha_t)}\,\Bigg| \geq \varepsilon\right)\,=\,\notag\\[0.5mm]
            &\sum_{k=0}^{n-2}\mathbb{P}\left( \Bigg|\,\frac{\tfrac{1}{2}\sum_{t=1}^{n}(1+\alpha_t)w_t}{\tfrac{1}{2}\sum_{t=1}^{n}(1+\alpha_t)}\,\Bigg| \geq \varepsilon\,\bigg|\,Z=k\right) \mathbb{P}(Z = k)\notag\\[0.5mm]
            & = \sum_{k=0}^{n-2}\mathbb{P}\left( \Bigg|\,\frac{\sum_{t\in I_{k+1}}w_t}{k+1}\,\Bigg| \geq \varepsilon\right) \mathbb{P}(Z = k),
        \end{align}
        where $I_{k+1}$ is a random index set of length $k+1$. Note, that the index sets of length $k+1$ are uniformly distributed, since every one of them has the same probability, $1/2^{n-2}$. For every (finitely many) realization of $I_{k+1}$, the variance proxy of 
        $(\sum_{t\in I_{k+1}}w_t)/(k+1)$ 
        is always
        $\sigma\sqrt{1/(k+1)}$,
        which follows from the properties of subgaussian variables \cite[Lemma 5.4]{lattimore_szepesvari_2020}. Then, the above probability \eqref{equ:lemma2_first_part} can be upper bounded 
        by
        using the Hoeffding inequality \cite[Proposition 2.5]{wainwright_2019}
        \begin{align*}
            &\leq\, 2\sum_{k=0}^{n-2}\mathbb{P}(Z = k)\exp\!\left( -\frac{(k+1)\varepsilon^2}{2 \sigma^2} \right)\\[0.5mm]
            &=\, 2\;\mathbb{E}\!\left[\exp\!\left( -\frac{Z \varepsilon^2}{2 \sigma^2} \right)\right]\exp\!\left( -\frac{\varepsilon^2}{2 \sigma^2} \right)\\[0.5mm]
            &=\, 2\left(\frac{1}{2}+\frac{1}{2}\exp\left(-\frac{\varepsilon^2}{2\sigma^2} \right) \right)^{\!n-2}\! \exp\!\left( -\frac{\varepsilon^2}{2 \sigma^2} \right)\\[0.5mm]
            &\leq\, 2\left(\frac{1}{2}+\frac{1}{2}\exp\left(-\frac{\varepsilon^2}{2\sigma^2} \right) \right)^{\!n-1}\hspace{-5mm},
        \end{align*}
        where 
        we applied the 
        moment generating function of the binomial distribution, to express $\mathbb{E}[\hspace{0.3mm}\exp(t\hspace{0.3mm}Z)\hspace{0.3mm}]$. $\square$
    \end{pf}

	Now, we state our nonasymptotic high probability upper bound for the length of SPS confidence intervals.
	\smallskip
	\begin{thm}\label{thm:param_iden_sample_complex}
		{\em Assuming A1, A3, A4, and considering system \eqref{equ:param_idenfy_sys}, the confidence intervals generated by SPS are shrinking at a geometric rate, i.e., for all $\varepsilon > 0$,}\vspace{1mm}
		\begin{align}\label{equ:param_iden_bound}
				&\mathbb{P}\Big(\sup_{\theta \in \confreg[p]}\abs{\hspace{0.3mm}\theta - \theta^*} \geq \varepsilon\Big) \leq \notag	\\
                &4\,\hspace{0.3mm}(m-q)\left(\frac{1}{2}+\frac{1}{2}\exp\left(-\frac{\varepsilon^2}{2\sigma^2} \right) \right)^{\!n-1}\hspace{-5mm}.
		\end{align}
	\end{thm}
    \vspace{-2mm}
	\begin{pf}
		First, we derive the sample complexity of SPS for the case where there are only two sums $S_0(\theta)$ and $S_1(\theta)$, and later generalize this to the case, where there are $m$ sums. In the case of two sums, $S_0(\theta)$ and $S_1(\theta)$ are:\vspace{0.5mm}
		\begin{align}
			S_0(\theta) &\,=\, n(\theta^* - \theta) + w,\\[1mm]
			S_1(\theta) &\, =\, \alpha(\theta^* - \theta) + w_{\alpha},
			\vspace{1mm}
		\end{align}
		where $n$ is the sample size, $w \defeq \sum_{t=1}^{n}w_t$, $\alpha \defeq \sum_{t=1}^{n}\alpha_t$ and $w_{\alpha} \defeq \sum_{t=1}^{n}\alpha_tw_t$. Then, a $p=0.5$ confidence set is
		\vspace{1mm}
		\begin{equation}
		    \label{confint05}
			\confreg[0.5] = \{\,\theta: S_0^2(\theta) \leq S_1^2(\theta)\, \} = \{\,\theta: \theta_1 \leq \theta \leq \theta_2 \,\},
		\vspace{1mm}			
		\end{equation}
		where $\theta_1$ and $\theta_2$ are the intersections of the two parabolas determined by $S_0^2(\theta)$ and $S_1^2(\theta)$. If the confidence region is finite, then $\theta_1$ and $\theta_2$ always exist. The pair $\theta_1$ and $\theta_2$ can be calculated by solving the quadratic equation
		\vspace{1mm}		
		\begin{equation}
			S_0^2(\theta) - S_1^2(\theta) = a^2\theta^2 + 2b \theta + c = 0,
		\vspace{1mm}			
		\end{equation}
		where
			$a = (n^2 - \alpha^2)$,
			$b = (\alpha^2-n^2)\theta^* + \alpha w_{\alpha} - nw$
		and
			$c = (\theta^*)^2(n^2 - \alpha^2) + 2\theta^*(nw - \alpha w_{\alpha}) + w^2 - w_{\alpha}^2$.
		The solutions are\vspace{-2mm}
		\begin{align}
				&\theta_{1} = \dfrac{w + w_{\alpha} + \alpha\theta^* + n\theta^*}{\alpha + n} = \dfrac{w + w_{\alpha}}{\alpha + n} + \theta^* = \notag\\
				&\dfrac{\sum_{t=1}^{n}w_t(1 + \alpha_t)}{\sum_{t=1}^{n}(1 + \alpha_t)} + \theta^*\!\!,
		\end{align}
		\begin{align}
				&\theta_{2} = \dfrac{-w + w_{\alpha} + \alpha\theta^* - n\theta^*}{\alpha - n} = \dfrac{w - w_{\alpha}}{n - \alpha} + \theta^* = \notag\\
				&\dfrac{\sum_{t=1}^{n}w_t(1 - \alpha_t)}{\sum_{t=1}^{n}(1 - \alpha_t)} + \theta^*\!\!.
		\end{align}
		Throughout our analysis we give concentration inequalities for the quantities $\abs{\hspace{0.3mm}\theta_1 - \mathbb{E}[\theta_1]}$ and $\lvert \hspace{0.3mm} \theta_2 - \mathbb{E}[\theta_2]\rvert$. The sequences $\{w_t\}$ and $\{\alpha_t\}$ are independent, and $\{w_t\}$ is centered, thus $\mathbb{E}[\theta_1] = \mathbb{E}[\theta_2] = \theta^*$.
		Using the results from Lemma \ref{lemma:intersection_is_sg},
        it holds for for $j = 1,2$ and any $\varepsilon > 0$ that
        \vspace{-0.5mm}
		\begin{equation}
			\mathbb{P}(\abs{\hspace{0.3mm}\theta_j - \theta^*} \geq \varepsilon) \,\leq\, 
        2 \left(\frac{1}{2}+\frac{1}{2}\exp\left(-\frac{\varepsilon^2}{2\sigma^2} \right) \right)^{\!n-1}\hspace{-5mm}.
		\end{equation}	
		Then, using the union bound (Boole's inequality), we have
		\begin{equation*}
			\mathbb{P}\Big(\sup_{\theta \in \confreg[0.5]}\abs{\hspace{0.3mm}\theta - \theta^*} \geq \varepsilon\Big) =\, 
			\mathbb{P}\Big(\max_{j=1,2}\abs{\hspace{0.3mm}\theta_j - \theta^*} \geq \varepsilon\Big) \,\leq
		\end{equation*}	
		\begin{equation}
		    \label{prob-bound-for-i}
            4 \left(\frac{1}{2}+\frac{1}{2}\exp\left(-\frac{\varepsilon^2}{2\sigma^2} \right) \right)^{\!n-1}\hspace{-5mm},
		    \vspace{1mm}
		\end{equation}	
		for all $\varepsilon > 0$. Note that \eqref{prob-bound-for-i} only holds for the special case of $m=2$ and $q=1$ (i.e., confidence probability $0.5$).

		Now, we consider the general case, i.e., we allow arbitrary $0 < q < m$ (integer) choices. Our aim will be to provide an upper bound for the probability of the ``bad'' event that the length of the constructed interval is above a given $\varepsilon > 0$. First, we can construct bad events for each $i$, i.e., the event that the $0.5$ probability interval defined by $S_0(\theta)$ and $S_i(\theta)$, cf.\ \eqref{confint05}, has length at least $\varepsilon$. This event is
		\vspace{1mm}
		\begin{equation}
		B_i^{\hspace{0.3mm}\varepsilon}\, \defeq \,\big\{\,\omega \in \Omega : \max_{j=1,2}\abs{\hspace{0.3mm}\theta_{i,j}(\omega) - \theta^*} \geq \varepsilon \, \big\},
		\vspace{1mm}
		\end{equation}
		where $\theta_{i,1}$ and $\theta_{i,2}$ are the two intersections of parabolas $S^2_0(\theta)$ and $S^2_i(\theta)$; and $\Omega$ is the sample space of the underlying probability space $(\Omega, \mathcal{F}, \mathbb{P})$. We already provided an upper bound for $\mathbb{P}(B_i^{\hspace{0.3mm}\varepsilon})$, see \eqref{prob-bound-for-i}, which is valid for all $i = 1, \dots, m-1$; but $B_1^{\hspace{0.3mm}\varepsilon}, \dots, B_{m-1}^{\hspace{0.3mm}\varepsilon}$ are not independent.
         
        By using the construction of SPS, see \citep{Csaji2015}, the ``good'' event, given integers $0 < q < m$, is
        \vspace{0.5mm}
		\begin{equation}
		G^{\hspace{0.3mm}\varepsilon}\, \defeq\hspace{-1mm} \bigcup_{\substack{I \subseteq \mathcal{M},\\ |I| \geq m-q}}\hspace{-1mm} \bigcap_{i \in I} \,G_i^{\hspace{0.3mm}\varepsilon},
		\end{equation}
		where $\mathcal{M} \defeq \{1,\dots, m-1\}$ and $G_i^{\hspace{0.3mm}\varepsilon} = \Omega \setminus B_i^{\hspace{0.3mm}\varepsilon}$, for $i\in\mathcal{M}$.
		
		This means that there exist at least $m-q$ (perturbed) parabolas, $\{S^2_i\}_{i \neq 0}$, such that all of their intersections with the reference $S^2_0$ are closer to $\theta^*$ than the given $\varepsilon > 0$.
		
		Then, by using De Morgan's laws, the  ``bad'' event is
        \vspace{0.5mm}
		\begin{equation}
		B^{\hspace{0.3mm}\varepsilon}\, \defeq \; \Omega \setminus G^{\hspace{0.3mm}\varepsilon} \,=\hspace{-1mm} \bigcap_{\substack{I \subseteq \mathcal{M},\\ |I| \geq m-q}}\hspace{-1mm} \bigcup_{i \in I} \,B_i^{\hspace{0.3mm}\varepsilon} \,=\hspace{-1mm} \bigcap_{\substack{I \subseteq \mathcal{M},\\ |I| = m-q}}\hspace{-1mm} \bigcup_{i \in I} \,B_i^{\hspace{0.3mm}\varepsilon}.
		\vspace{-2mm}
		\end{equation}
		
		The probability of this ``bad'' event can be bounded by
		\begin{equation}
		\mathbb{P}(B^{\hspace{0.3mm}\varepsilon})\, \leq \min_{\substack{I \subseteq \mathcal{M},\\ |I| = m-q}}\hspace{-2mm} \mathbb{P}\!\left[\,\bigcup_{i \in I}\,B_i^{\hspace{0.3mm}\varepsilon} \right] \, \leq\, (m-q)\cdot \mathbb{P}(B_1^{\hspace{0.3mm}\varepsilon}),
		\end{equation}
		where we used that $\mathbb{P}(A \cap B) \leq \min\{\mathbb{P}(A),\, \mathbb{P}(B)\}$ and $\mathbb{P}(A \cup B) \leq \mathbb{P}(A) + \mathbb{P}(B)$. This completes the proof. $\square$

	\end{pf}
	Next, we give a high probability bound on the size of the outer approximation of the SPS confidence interval. 
	\begin{cor}\label{cor:param_iden_sample_complex_oa}
		{\em Assuming A1, A3 and A4, and considering \eqref{equ:param_idenfy_sys}, the confidence region of the SPS outer approximation is shrinking at a geometric rate, i.e., for all} $\varepsilon > 0$,
		\begin{align}
			&\mathbb{P}\Big(\sup_{\theta \in \outappr[p]}|\hspace{0.3mm}\theta - \hat{\theta}_n| \geq \varepsilon\Big)  \,\leq\, \notag\\
			&4\,\hspace{0.3mm}(m-q)\left(\frac{1}{2}+\frac{1}{2}\exp\left(-\frac{\varepsilon^2}{8\sigma^2} \right) \right)^{\!n-1}\hspace{-5mm}.
		\end{align}
	\end{cor}
    \vspace{-2mm}
	\begin{pf}
		The proof of this corollary is similar to the proof of Theorem \ref{thm:param_iden_sample_complex}, we first consider the case for two sums: $S_0(\theta)$ and $S_1(\theta)$, and then generalize it to arbitrary $0 < q < m$ choices. For two sums the size of the SPS confidence region induced outer approximation is $2\max(|\hspace{0.3mm}\theta_{1} - \hat{\theta}_n|, |\hspace{0.3mm}\theta_{2} - \hat{\theta}_n|)$, 
		therefore, we should study $\theta_{1} - \hat{\theta}_n$.
		\begin{align}
			&\theta_{1} - \hat{\theta}_n = \dfrac{\sum_{t=1}^{n}w_t(1 + \alpha_t)}{\sum_{t=1}^{n}(1 + \alpha_t)} + \theta^* - \theta^* - \dfrac{\sum_{t=1}^{n}w_t}{n}\notag\\[1mm]
			&\overset{d}{=}\; \dfrac{\sum_{t=1}^{n}w_t(1 + \alpha_t)}{\sum_{t=1}^{n}(1 + \alpha_t)} + \dfrac{\sum_{t=1}^{n}w_t}{n}.
		\end{align}
        By introducing $Z = \sum_{t=3}^{n}(1+\alpha_t)/2$ and using the same ideas and 
        notations as in the proof of Lemma \ref{lemma:intersection_is_sg}, we have
        \begin{align}\label{equ:proof_oa}
            &\mathbb{P}\left( \Bigg|\,\frac{\sum_{t=1}^{n}w_t(1 + \alpha_t)}{\sum_{t=1}^{n}(1 + \alpha_t)} + \frac{\sum_{t=1}^{n}w_t}{n}\,\Bigg| \geq \varepsilon\right)=\notag\\
            &\sum_{k=0}^{n-2}\mathbb{P}\left( \Bigg|\,\frac{\tfrac{1}{2}\sum_{t=1}^{n}w_t(1 + \alpha_t)}{\tfrac{1}{2}\sum_{t=1}^{n}(1 + \alpha_t)} + \frac{\sum_{t=1}^{n}w_t}{n}\,\Bigg| \geq \varepsilon\,\bigg|\,Z=k\right) \notag\\
            &\cdot\mathbb{P}(Z = k) = \sum_{k=0}^{n-2}\mathbb{P}\Bigg( \Bigg|\,\sum_{t\in I_{k+1}}w_t\,\left(\frac{1}{k+1} + \frac{1}{n}\right) + \notag\\
            &\frac{1}{n}\sum_{t\notin I_{k+1}}w_t\,\Bigg|\geq \varepsilon\Bigg) \mathbb{P}(Z = k)\leq 2\sum_{k=0}^{n-2}\mathbb{P}(Z = k)\notag\\
            &\cdot\exp\left( \frac{-\varepsilon^2}{2 \sigma^2\left[(k+1)\left(\frac{1}{k+1} + \frac{1}{n}\right)^2+(n-k-1)\left(\frac{1}{n}\right)^2\right]} \right)\notag\\
            & = 2\sum_{k=0}^{n-2}\mathbb{P}(Z = k)\exp\left( -\frac{\varepsilon^2}{2 \sigma^2\left(\frac{1}{k+1} + \frac{3}{n}\right)}\right)\notag\\
            & = 2\sum_{k=0}^{n-2}\mathbb{P}(Z = k)\exp\left( -\frac{\varepsilon^2n(k+1)}{2 \sigma^2\left(n + 3(k+1)\right)}\right)\notag\\
            &\,\leq\, 2\sum_{k=0}^{n-2}\mathbb{P}(Z = k)\exp\left( -\frac{\varepsilon^2(k+1)}{8 \sigma^2}\right)\notag\\ 
            &\,\leq\, 2\left(\frac{1}{2}+\frac{1}{2}\exp\left(-\frac{\varepsilon^2}{8\sigma^2} \right) \right)^{\!n-1}\hspace{-5mm},
        \end{align}
        where we also expoited that for every (finitely many) realization of $I_{k+1}$ the sum $\sum_{t\in I_{k+1}}w_t\,\left(1/(k+1) + 1/n\right) + (1/n) \sum_{t\notin I_{k+1}}w_t$ has a (common) variance proxy
        \begin{align*}
            \sigma\left[(k+1)\left(\frac{1}{k+1} + \frac{1}{n}\right)^2+(n-k-1)\left(\frac{1}{n}\right)^2\right]^{1/2}\hspace{-4.5mm}.
        \end{align*}

		Then, using the union bound %
		\vspace*{-0.5mm}
        \begin{align}		
		\mathbb{P}\Big(\max_{j=1,2}\abs{\hspace{0.3mm}\theta_j - \hat{\theta}_n} \geq \varepsilon\Big) \leq\, 4\left(\frac{1}{2}+\frac{1}{2}\exp\left(-\frac{\varepsilon^2}{8\sigma^2} \right) \right)^{n-1}\hspace{-5.5mm}.\notag
	    \end{align}		
		Finally, the above result, which assumed $m=2$ and $q=1$, can be generalized to arbitrary choices of $0 < q < m$ in the same way as we did in the proof of Theorem \ref{thm:param_iden_sample_complex}. $\square$
	\end{pf}
	
	\vspace{-3mm}
	\subsection{Scalar linear regression with bounded regressors}
	\vspace{-2mm}
	Consider the following scalar linear regression problem
	\vspace{1mm}
	\begin{equation}\label{equ:one_dim_lr_sys}
		y_t = \varphi_t\hspace{0.3mm}\theta^* + w_t,
		\vspace{1mm}
	\end{equation}
	where $\varphi_t$ the regressor, $y_t$ is the output, $w_t$ is the noise and $\theta^*$ is the true parameter to be estimated. Both the random variables $\varphi_t$, $y_t$, $w_t$, and the true parameter $\theta^*$ are scalars.
	Our assumption on the noise is still A3, and we make the following assumption on the regressors:
	\begin{itemize}
		\item[A5] {\em The regressor sequence, $\{\varphi_t\}$, consists of independent random variables that are almost surely bounded from below: for all $t$, we {\em(}a.s.{\em)} have $0 < \varphi_{\text{min}} \leq \abs{\hspace{0.3mm}\varphi_t}$.}
	\end{itemize}
	Similarly to the previous identification case, we first state and prove a lemma, then two concentration inequalities regarding the size of the confidence region generated by the SPS algorithm and its outer approximation.
	\smallskip
    \begin{lem}\label{lemma:intersection_is_sg_lr}
		{\em Assume A1, A2, A3 and A5, $n > 2$, and let\vspace{1mm}
		$$X_{1,2} \defeq \frac{\sum_{t=1}^{n}\varphi_t(1\pm\alpha_t)w_t}{\sum_{t=1}^{n}\varphi_t^2(1\pm\alpha_t)}$$
        where $\alpha_1 = 1$, $\alpha_2 = -1$, and $\{\alpha_t\}_{t=3}^n$ are i.i.d.\ Rademacher random variables, independent of the noise terms $\{w_t\}$.
        
        Then, variables $X_1$ and $X_2$ satisfy for all $\varepsilon \geq 0:$
        \begin{equation}
        \mathbb{P}\left( |X_{1,2}| \geq \varepsilon\right) \, \leq\, 2\left(\frac{1}{2}+\frac{1}{2}\exp\left(-\frac{\varepsilon^2\varphi_{\text{min}}^2}{2\sigma^2} \right) \right)^{\!n-1}\hspace{-5mm}.
        \vspace{-2mm}
        \end{equation}}
	\end{lem}
	\begin{pf}
        Note that as $X_{1}\overset{d}{=}X_{2}$, it is enough to prove the claim for $X_{1}$. Then, by introducing $Z = \sum_{t=3}^{n}(1+\alpha_t)/2$ and using the same arguments and notations as in the proof of Lemma \ref{lemma:intersection_is_sg}, an overbound can be calulated as
        \begin{align}
            &\mathbb{P}\left( |X_1| \geq \varepsilon\right) \,=\, \mathbb{P}\left( \Bigg|\,\frac{\sum_{t=1}^{n}\varphi_t(1+\alpha_t)w_t}{\sum_{t=1}^{n}\varphi_t^2(1+\alpha_t)}\,\Bigg| \geq \varepsilon\right)\,=\notag\\
            &\sum_{k=0}^{n-2}\mathbb{P}\left( \Bigg|\,\frac{\tfrac{1}{2}\sum_{t=1}^{n}\varphi_t(1+\alpha_t)w_t}{\tfrac{1}{2}\sum_{t=1}^{n}\varphi_t^2(1+\alpha_t)}\,\Bigg| \geq \varepsilon\,\bigg|\,Z=k\right) \mathbb{P}(Z = k)\notag\\
            & = \sum_{k=0}^{n-2}\mathbb{P}\left( \Bigg|\,\frac{\sum_{t\in I_{k+1}}\varphi_tw_t}{\sum_{t\in I_{k+1}}\varphi_t^2}\Bigg| \geq \varepsilon\right) \mathbb{P}(Z = k)\notag\\
            &\,\leq\, 2\sum_{k=0}^{n-2}\mathbb{P}(Z = k)\exp\!\left( -\frac{\varepsilon^2(k+1)\varphi_{\text{min}}^2}{2\sigma^2} \right)\notag\\
            &\,\leq\, 2\left(\frac{1}{2}+\frac{1}{2}\exp\left(-\frac{\varepsilon^2\varphi_{\text{min}}^2}{2\sigma^2} \right) \right)^{\!n-1}\hspace{-5mm},
        \end{align}
        where we used that for every realization of the random index set $I_{k+1}$ and inputs $\{\varphi_t\}$ we can choose a common (realization independent) variance proxy, $
        \sigma_{\text{max}}$, for the variable $(\sum_{t\in I_{k+1}}\varphi_tw_t)/(\sum_{t\in I_{k+1}}\varphi_t^2)$, since 
        \vspace{-1.5mm}
        \begin{align*}
            \sigma_{\text{max}}\, \defeq\, \sigma\left(\frac{1}{(k+1)\varphi_{\text{min}}^2}\right)^{\!1/2} \geq\, \sigma\left(\frac{\sum_{t\in I_{k+1}}\varphi_t^2}{\left(\sum_{t\in I_{k+1}}\varphi_t^2\right)^2}\right)^{\!\!1/2}\hspace{-4mm},
        \end{align*}
        where the right hand side is the (conditional) variance proxy of $X$, given $\{Z=k\}$, $I_{k+1}$, and $\{\varphi_t\}$. $\Box$
	\end{pf}
	\smallskip
	\begin{thm}\label{thm:bounded_lr_sample_complex}
		{\em Assuming A1, A2, A3, A4 and A5, and considering system \eqref{equ:one_dim_lr_sys}, the confidence region generated by SPS is shrinking at a geometric rate, i.e, for all $\varepsilon > 0$,}
		\begin{align}
            &\mathbb{P}\Big(\sup_{\theta \in \confreg[p]}\abs{\hspace{0.3mm}\theta - \theta^*} \geq \varepsilon\Big) \leq\notag\\
            &4\,\hspace{0.3mm}(m-q)\left(\frac{1}{2}+\frac{1}{2}\exp\left(-\frac{\varepsilon^2\varphi_{\text{min}}^2}{2\sigma^2} \right) \right)^{\!n-1}\hspace{-5mm}.
		\end{align}
	\end{thm}
    \vspace{-2mm} 
	\begin{pf}
		The proof is similar to that of Theorem \ref{thm:param_iden_sample_complex}, we also investigate $\lvert\theta_{1} - \mathbb{E}(\theta_{1})\rvert$ and $\lvert\theta_{2} - \mathbb{E}(\theta_{2})\rvert$,  for system \eqref{equ:one_dim_lr_sys}, in the case of two sums. The points of intersection can be calculated by solving the 
        equation $S_0^2(\theta)-S_1^2(\theta) = 0$ as in the proof of Theorem \ref{thm:param_iden_sample_complex}. The two solutions are
		\begin{equation}
				\theta_{1} = \dfrac{\sum_{t=1}^{n}\varphi_tw_t(1 + \alpha_t)}{\sum_{t=1}^{n}\varphi_t^2(1 + \alpha_t)} + \theta^*,
		\end{equation}
		\begin{equation}
				\theta_{2} = \dfrac{\sum_{t=1}^{n}\varphi_tw_t(1 - \alpha_t)}{\sum_{t=1}^{n}\varphi_t^2(1 - \alpha_t)} + \theta^*.
		\end{equation}
		As in the ''constant in noise'' identification case, we can apply the concentration bound of Lemma \ref{lemma:intersection_is_sg_lr} for $\lvert\theta_{1} - \mathbb{E}(\theta_{1})\rvert$ and $\lvert\theta_{2} - \mathbb{E}(\theta_{2})\rvert$. The case of more than two sums 
        can be constructed the same way as in the proof of Theorem \ref{thm:param_iden_sample_complex}. $\square$
	\end{pf}
	\smallskip
	\begin{cor}\label{thm:bounded_lr_sample_complex_oa}
		{\em Assuming A1-A5, and considering system \eqref{equ:one_dim_lr_sys}, the outer approximations of the SPS confidence regions are shrinking at a geometric rate: for all $\varepsilon > 0$,}
		\begin{align}
		&\mathbb{P}\Big(\sup_{\theta \in \outappr[p]}|\hspace{0.3mm}\theta - \hat{\theta}_n| \geq \varepsilon\Big)  \leq \notag \\
        &4\,\hspace{0.3mm}(m-q)\left(\frac{1}{2}+\frac{1}{2}\exp\left(-\frac{\varepsilon^2\varphi_{\text{min}}^2}{8\sigma^2} \right) \right)^{\!n-1}\hspace{-5mm}.
		\end{align}
	\end{cor}
    \vspace{-1mm}
	\begin{pf}
        We only provide a proof sketch, since the proof is very similar to the proof of Corollary \ref{cor:param_iden_sample_complex_oa}. The difference is that the term under investigation is
        \begin{align}
            \abs{\frac{\sum_{t=1}^{n}\varphi_t(1\pm\alpha_t)w_t}{\sum_{t=1}^{n}\varphi_t^2(1\pm\alpha_t)} + \frac{\sum_{t=1}^{n}\varphi_tw_t}{\sum_{t=1}^{n}\varphi_t^2}}
        \end{align}
        By applying the the same ideas as we did in the proof of Corollary \ref{cor:param_iden_sample_complex_oa}, namely, the law of total probability, constructing the random index set $I_{k+1}$, deriving the variance proxy of the sum for every realization of $I_{k+1}$, $\{\varphi_t\}$ and the Hoeffding inequality, it can be derived that
        \begin{align}
            &\mathbb{P}\Big(\sup_{\theta \in \outappr[p]}|\hspace{0.3mm}\theta - \hat{\theta}_n| \geq \varepsilon\Big)  \leq\\
            &2\sum_{k=0}^{n-2}\mathbb{P}(Z = k)\exp\left( -\frac{\varepsilon^2\varphi_{\text{min}}^2}{2 \sigma^2\left(\frac{1}{k+1} + \frac{3}{n}\right)}\right)\notag
        \end{align}
        Using last two steps of \eqref{equ:proof_oa} completes the proof. $\square$
        
	\end{pf}
	\subsection{Scalar linear regression with unbounded regressors}
	\vspace{-2mm}
	Consider the scalar linear regression problem \eqref{equ:one_dim_lr_sys} and the following filtration $\mathbb{F} = \{\mathcal{F}_t\}_{t\geq0}$, where
	\begin{equation}
		\mathcal{F}_t\, \defeq\, \sigma\{w_1, \dots, w_t, \alpha_1, \dots, \alpha_{t+1}, \varphi_1, \dots, \varphi_{t+1} \}.
	\end{equation}
	Our assumptions on the noises and on the regressors are
	\begin{itemize}
		\item[A6] {\em Let $\{w_t\}$ be a sequence of independent, homoscedastic, conditionally $\sigma$-subgaussian random variables with variance $\sigma^2$, furthermore, 
        for all $t$, let $w_{t+1}$ be independent of $\mathcal{F}_t$. Formally, $\forall\, \lambda \in \mathbb{R}$ and $t\, \geq\, 1:$}
		\begin{equation}
			\mathbb{E}\hspace{-0.3mm}\left[\hspace{0.3mm}\exp\left(\lambda w_t\right)\vert\mathcal{F}_{t-1}\hspace{0.3mm}\right] \,\leq\, \exp\left(\dfrac{\lambda^2\sigma^2}{2}\right),
		\end{equation}
		\begin{equation}
			\mathbb{E}\hspace{-0.3mm}\left[\hspace{0.3mm}w_{t+1}\vert\mathcal{F}_{t}\hspace{0.3mm}\right]\, =\, \mathbb{E}\left[w_{t+1}\right]= 0.
			\vspace{2mm}
		\end{equation}
		\item[A7] {\em Let $\{\varphi_t\}$ be a sequence of identically distributed random variables that are %
        integrable.}
	\end{itemize}
	In order to give a high probability bound on the size of the confidence region constructed by SPS in this unbounded regressor case, we first show in Lemma \ref{lemma:Mn_is_martingale} and Lemma \ref{lemma:Mn_is_sg_martingale} that $M_n = \sum_{t=1}^{n}\varphi_tw_t(1+\alpha_t)$ is a subgaussian martingale with respect to the filtration $\mathbb{F}$.
	\begin{lem}\label{lemma:Mn_is_martingale}
		{\em Assuming A4, A6 and A7, the sum $M_n=\\\sum_{t=1}^{n}\varphi_tw_t(1+\alpha_t)$ is a martingale w.r.t.\ filtration $\mathbb{F}$.}
	\end{lem}
	\begin{pf}
	    We need to check two properties:
		\begin{align}
				&\mathbb{E}\abs{M_n} = \mathbb{E}\!\left[\abs{\sum_{t=1}^{n}\varphi_tw_t(1+\alpha_t)}\right] \leq \mathbb{E}\!\left[\sum_{t=1}^{n}\abs{\varphi_tw_t(1+\alpha_t)}\right] \notag\\
				& \leq\sum_{t=1}^{n}\mathbb{E}\abs{\varphi_t}\mathbb{E}\abs{w_t}\mathbb{E}\abs{(1+\alpha_t)} < \infty,
		\end{align}
		since $\{\varphi_t\}$ are integrable, $\{w_t\}$ are subgaussian and $\{\alpha_t\}$ are Rademacher random variables.      
		\begin{align}
				&\mathbb{E}\left[M_{n+1}\vert\mathcal{F}_t\right] = \mathbb{E}\left[M_{n}\vert\mathcal{F}_t\right] + \mathbb{E}\left[\varphi_{n+1}w_{n+1}(1+\alpha_{n+1})\vert\mathcal{F}_t\right]\notag\\
				& = M_n,
		\end{align}
        which completes the proof. $\square$
	\end{pf}
	The martingale $\{X_n\}$ is called subgaussian \citep{bercu2008}, if $\exists\, \beta > 0$, such that for $\forall\, n \geq 1$, $\lambda \in \mathbb{R}$:
	\begin{equation}
		\mathbb{E}\left[\hspace{0.3mm}\exp(\lambda\Delta X_n)\,\vert\,\mathcal{F}_{n-1}\hspace{0.3mm}\right]\, \leq\, \exp\left(\dfrac{\beta^2\lambda^2}{2}\Delta \langle X\rangle_n\right)\hspace{-0.3mm},
	\end{equation}
	where $\langle X\rangle_n$ is the quadratic variation of the martingale $X_n$. The quadratic variation of $M_n$ can be written as
    \vspace{-0.5mm}
	\begin{align}
			&\langle M\rangle_n = \sum_{t=1}^{n}\mathbb{E}\left[(M_{t}-M_{t-1})^2\vert\mathcal{F}_{t-1}\right] = \notag\\[-1mm]
			&\sum_{t=1}^{n}\mathbb{E}\left[\varphi_t^2w_t^2(1+\alpha_t)^2\vert\mathcal{F}_{t-1}\right] = \sum_{t=1}^{n}\varphi_t^2\sigma^2(1+\alpha_t)^2 = \notag\\[-1mm]
			&2\,\sigma^2\sum_{t=1}^{n}\varphi_t^2(1+\alpha_t).
	\end{align}
	\begin{lem}\label{lemma:Mn_is_sg_martingale}
		{\em Assuming A4, A6 and A7 the martingale $\{M_n\}$ is subgaussian with $\beta = 1$.}
	\end{lem}
	\begin{pf}
		\begin{align}
				& \mathbb{E}\left[\exp(\lambda\Delta M_n)\vert\mathcal{F}_{n-1}\right] = \mathbb{E}\left[\exp(\lambda\varphi_nw_n(1+\alpha_n))\vert\mathcal{F}_{n-1}\right]\notag\\
				& \leq \exp\left(\dfrac{\lambda^2\varphi_n^2\sigma^2(1+\alpha_n)^2}{2}\right) = \exp\left(\lambda^2\varphi_n^2\sigma^2(1+\alpha_n)\right) \notag\\
				& = \exp\left(\dfrac{\lambda^2}{2}\Delta \langle M\rangle_n\right),
		\end{align}
		therefore $\beta = 1$. $\square$
	\end{pf}
	In the following we state and prove our theorem by using the fact that the bounds of the confidence region can be written as self-normalized martingales (w.r.t.\ $\mathbb{F}$). To give a concentration inequality for the self-normalized martingale we use the result from \citep{bercu2008}.
	\begin{thm}\label{thm:ubounded_lr_sample_complex}
		{\em Assuming A1, A2, A4, A6 and A7, and considering system \eqref{equ:one_dim_lr_sys}, the SPS confidence sets are shrinking according to the following inequality, for all $\varepsilon > 0$},
		\begin{align}
				&\mathbb{P}\Big(\sup_{\theta \in \confreg[p]}\abs{\hspace{0.3mm}\theta - \theta^*} \geq \varepsilon\Big) \leq\\ &4\,(m-q)\left(\dfrac{1}{2} + \dfrac{1}{2}\mathbb{E}\left[\exp\left(-\dfrac{\varepsilon^2\varphi_0^2}{\sigma^2}\right)\right]\right)^{\!\!\frac{n-1}{4}}\hspace{-5mm}.
		\end{align}
	\end{thm}
    \vspace{-2mm}	
    \begin{pf}
		As in the proof of Theorem \ref{thm:bounded_lr_sample_complex} we first investigate the distance between the intersection given by the parabolas and true parameter in the case of two sums
		\begin{equation}
			\theta_1 - \theta^* = \dfrac{\sum_{t=1}^{n}\varphi_tw_t(1 + \alpha_t)}{\sum_{t=1}^{n}\varphi_t^2(1 + \alpha_t)} = \dfrac{2\sigma^2M_n}{\langle M\rangle_n},
		\end{equation}
		with respect to the filtration $\mathbb{F}$.
		For a subgaussian martingale $X_n$ the following concentration inequality holds \citep[formula (4.6)]{bercu2008}
		\vspace{-1mm}
		\begin{align}
				&\mathbb{P}\left(\dfrac{X_n}{a+b\langle X\rangle_n} \geq \varepsilon\right) \leq \notag\\
				&\inf_{p > 1}\left(\mathbb{E}\!\left[\exp\left(-(p-1)\dfrac{\varepsilon^2}{\beta^2}\left(ab+\dfrac{b^2}{2}\langle X\rangle_n\right)\right)\right]\right)^{\!1/p}\hspace{-4mm}.
		\end{align}
        Using this result, we have the following 
		\begin{align*}
				&\mathbb{P}\left(\theta_1 - \mathbb{E}\big[\theta_1\big]\geq \varepsilon\right) = \mathbb{P}\left(\dfrac{\sum_{t=1}^{n}\varphi_tw_t(1 + \alpha_t)}{\sum_{t=1}^{n}\varphi_t^2(1 + \alpha_t)}\geq \varepsilon\right)\notag\\
				& = \mathbb{P}\left(\dfrac{2\,\sigma^2M_n}{\langle M\rangle_n}\geq \varepsilon\right)\leq\notag\\
				&\inf_{p > 1}\left(\mathbb{E}\!\left[\exp\left(-(p-1)\frac{\varepsilon^22\sigma^2}{8\sigma^4}\left(\sum_{t=1}^{n}\varphi_t^2(1 + \alpha_t)\right)\right)\right]\right)^{\!1/p}\notag\\
				&= \inf_{p > 1}\left(\mathbb{E}\!\left[\prod_{t=1}^{n}\exp\left(-(p-1)\dfrac{\varepsilon^2}{4\sigma^2}\left(\varphi_t^2(1 + \alpha_t)\right)\right)\right]\right)^{\!1/p}\notag				
		\end{align*}
		\begin{align}
				&\leq \inf_{p > 1}\left(\prod_{t=1}^{n}\mathbb{E}\!\left[\exp\left(-2(p-1)\dfrac{\varepsilon^2}{4\sigma^2}\left(\varphi_t^2(1 + \alpha_t)\right)\right)\right]\right)^{\!1/2p}\notag\\
				&\leq \left(\prod_{t=1}^{n}\mathbb{E}\!\left[\exp\left(-\dfrac{\varepsilon^2}{2\sigma^2}\left(\varphi_t^2(1 + \alpha_t)\right)\right)\right]\right)^{1/4}\notag\\
				&= \left(\mathbb{E}\!\left[\exp\left(-\dfrac{\varepsilon^2}{2\sigma^2}\left(\varphi_0^2(1 + \alpha_0)\right)\right)\right]\right)^{\!\frac{n-2}{4}}\notag\\
				&\quad\;\cdot\,\mathbb{E}\!\left[\exp\left(-\dfrac{\varepsilon^2\varphi_0^2}{\sigma^2}\right)\right]^{\!\frac{1}{4}}\notag\\
				&\leq\left(\dfrac{1}{2} + \dfrac{1}{2}\mathbb{E}\left[\exp\left(-\dfrac{\varepsilon^2\varphi_0^2}{\sigma^2}\right)\right]\right)^{\!\frac{n-1}{4}}\hspace{-5mm},
		\end{align}
		where we used the Hölder inequality (line 4 to 5), the fact that $\{\varphi_t\}$ are identically distributed (line 6 to 7) and the law of total expectation (last step). In the derivations $\varphi_0$ is any random variable having the same distribution as $\{\varphi_t\}$ and $\alpha_0$ is a Rademacher variable.

        For the double-sided case, $\mathbb{P}\left(|\theta_1 - \mathbb{E}\big[\theta_1\big]|\geq \varepsilon\right)$, a $2$ multiplyer comes in, as before.
		The case of more than two sums
        can be handled the same way as in Theorem \ref{thm:param_iden_sample_complex}. $\square$

	\end{pf}
    Observe that if Var$(\varphi_t) > 0, (\forall t)$, then we have
	\begin{equation}
		\left(\dfrac{1}{2} + \dfrac{1}{2}\,\mathbb{E}\!\left[\exp\left(-\dfrac{\varepsilon^2\varphi_0^2}{\sigma^2}\right)\right]\right)\, <\, 1.
	\end{equation}
	To give an example of the high probability bound for a specific distribution, in Corollary \ref{cor:bound_for_normal} we investigate the case when the regressors are normally distributed.
	\smallskip
	\begin{cor}\label{cor:bound_for_normal}
		{\em Under the assumptions of Theorem \ref{thm:ubounded_lr_sample_complex}, and assuming that the regressors are sampled from a centered normal distribution, $\varphi_t \sim \mathcal{N}(0, \sigma_{\varphi}^2)$, the following 
		concentration inequality holds for the SPS confidence regions:}
		\begin{align}\label{equ:lr_bound_centered_normal}
				&\mathbb{P}\Big(\sup_{\theta \in \confreg[p]}\abs{\hspace{0.3mm}\theta - \theta^*} \geq \varepsilon\Big) \leq  \notag\\[-2mm]
				&4\,(m-q)\left(\dfrac{1}{2} +
				\dfrac{1}{2}\left(1+\dfrac{2\varepsilon^2\sigma_{\varphi}^2}{\sigma^2}\right)^{\!\!-\frac{1}{2}}\right)^{\!\!\!\frac{n-1}{4}}\hspace{-4.5mm}.\\[-7mm]
                \notag
		\end{align}
	\end{cor}
	\begin{pf}
		For variable $X \sim \mathcal{N}(\mu, \sigma_{X}^2)$ it holds that $\forall t < \frac{1}{2}$
		\begin{equation}
			\mathbb{E}[\exp(tX^2)] = \frac{1}{\sqrt{1-2t\sigma_{X}^2}}\exp\left(\frac{\mu^2t}{1-2t\sigma_{X}^2}\right).
		\end{equation}
		Since $-\tfrac{\varepsilon^2}{\sigma^2} < \tfrac{1}{2}$ it holds that
		\begin{equation}
			\mathbb{E}\!\left[\exp\left(-\dfrac{\varepsilon^2\varphi_0^2}{\sigma^2}\right)\right] = \frac{1}{\sqrt{1+\dfrac{2\varepsilon^2\sigma_{\varphi}^2}{\sigma^2}}}.\quad\square
		\end{equation}
	\end{pf}
	\section{Simulation experiments}\label{sec:simulation}
	\vspace{-2mm}
	In this section we compare our theoretical bounds on the size of the confidence region with the size of the region given by simulated trajectories. We consider the system,
	\begin{equation}
		y_t \,=\, \varphi_t\hspace{0.3mm}\theta^* + w_t,
	\end{equation}
	where $\theta^* = 5$ and $w_t \sim \text{Unif}(-1,1)$. Throughout our experiments we considered $0.5$-level confidence regions, that is $m = 2$ and $q=1$, a sample size of $n = 400$ and $k = 1000$ independently simulated trajectories. We present simulation results for the constant identification problem, where $\varphi_t = 1$, $\forall t$ and for the scalar linear regression with unbounded regressor case, where $\varphi_t \sim \mathcal{N}(0,1)$.
	\subsection{Constant identification}
\begin{figure}[ht]
		\begin{center}
			\includegraphics[width=8.4cm]{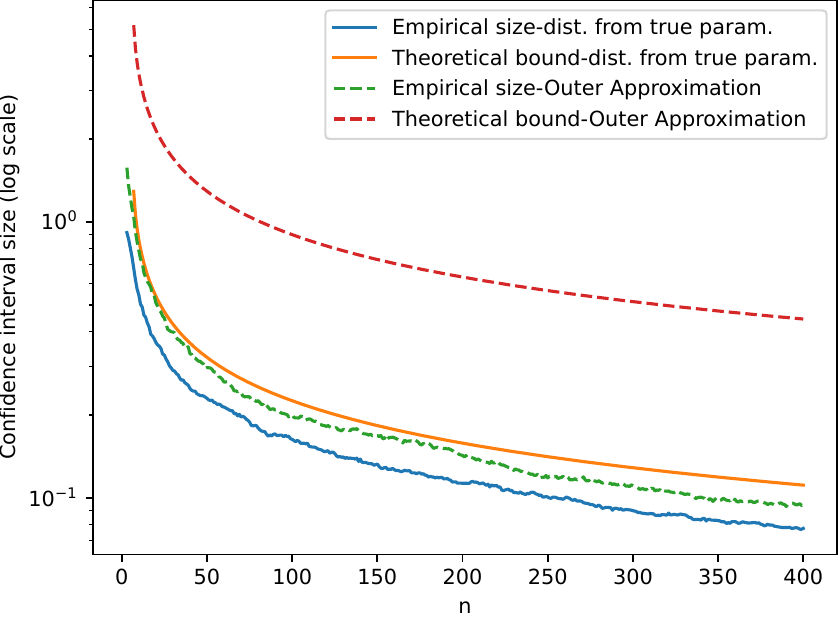}    %
			\caption{Comparison of the empirical size and the theoretical upper bound on the size in the constant identification case for $m=2, \delta=0.1, n=400, k =1000$.} 
			\label{fig:constant_iden_experiment}
		\end{center}
        \vspace{4mm}  
	\end{figure}

	\vspace{-2mm}
	The stochastic bound given in \eqref{equ:param_iden_bound} can be reformulated as 
    \begin{align*}
        \max_{i=1,2}\hspace{0.3mm}\abs{\hspace{0.3mm}\theta_i - \theta^*} \,\leq\, \sqrt{-2\hspace{0.5mm}\sigma^2\log\left(2\left(\frac{\delta}{4}\right)^{\frac{1}{n-1}}-1\right)},
    \end{align*}
	with probability at least $1- \delta$. Note that the above bound and the bound on the outer approximation below is only valid if $n > \tfrac{\ln(\delta/4)}{\ln(1/2)} + 1$.
 In our experiments we set $\delta = 0.1$ and from the noise setting it follows that $\sigma^2 = 1/3$ is the optimal variance proxy. We have compared the bound given above with $\max(\abs{\theta_{1,s} - \theta^*}, \abs{\theta_{2,s} - \theta^*})$, where $\theta_{1,s}$ and $\theta_{2,s}$ are calculated from the sample $\{(y_t, \varphi_t, w_t)\}_{t = 3}^n$. Note that at least 2 data is needed for the confidence region to be finite assuming $\alpha_1$ and $\alpha_2$ is different, therefore we set $\alpha_1 = 1$ and $\alpha_2 = -1$.
 The difference between the empirical size with confidence level $1-\delta$ and the theoretical size are shown in Fig. ~\ref{fig:constant_iden_experiment}. Empirical quantiles were used for each iteration, i.e., the smallest number for which at least the specified portion of simulation realizations are below that number.
 We also compared the theoretical bound and empirical size (quantiles) of the outer approximation of the SPS region.
 For the outer approximation it holds that
 \begin{align*}
     2\hspace{0.3mm}\max_{i=1,2}\hspace{0.3mm} |\hspace{0.3mm}\theta_i- \hat{\theta}_n|\, \leq\, \sqrt{-32\hspace{0.5mm}\sigma^2\log\left(2\left(\frac{\delta}{4}\right)^{\frac{1}{n-1}}-1\right)},
 \end{align*}
 with probability at least $1- \delta$.
 The empirical size of the outer approximation can be calculated from the sample $\{(y_t, \varphi_t)\}_{t = 3}^n$. Fig. ~\ref{fig:constant_iden_experiment}. shows the difference between the upper bound and the empirical size with confidence level $1-\delta$ for the outer approximation.
	\subsection{Scalar linear regression}
\begin{figure}[t]
		\begin{center}
			\includegraphics[width=8.4cm]{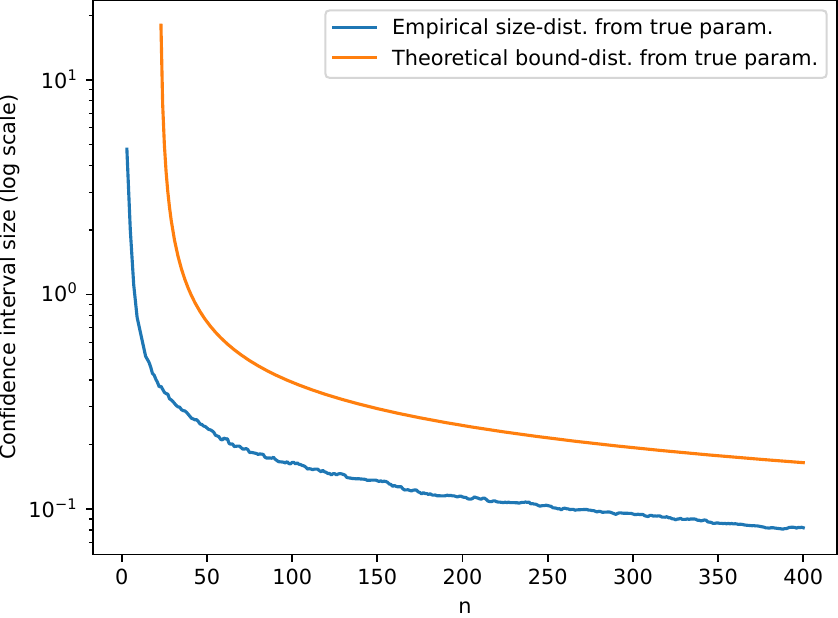}    %
			\caption{Comparison of the empirical size and the
            corresponding theoretical bound for scalar linear regression with Gaussian regressors, $m=2,  n=400, k =1000$.} 
			\label{fig:lr_experiment_indicator}
		\end{center}
        \vspace{4mm}
	\end{figure}

	\vspace{-2mm}
	The stochastic bound obtained in the centered normal regressor case \eqref{equ:lr_bound_centered_normal} can be reformulated as
	\begin{align*}
        \max_{i=1,2}\hspace{0.3mm}\abs{\hspace{0.3mm}\theta_i - \theta^*} \,\leq\, 
        \sqrt{\frac{\sigma^2\left(\left(\frac{1}{2\left(\frac{\delta}{4}\right)^{\frac{4}{n-1}}-1}\right)^{\!2}-1\right)}{2\hspace{0.5mm}\sigma_{\varphi}^2}},
	\end{align*}
	with probability at least $1- \delta$. Note that the above bound is only valid if $n > \tfrac{4\ln(\delta/4)}{\ln(1/2)} + 1$. We performed experiments with $\delta = 0.1$, from the experimental setting it follows that $\sigma^2 = 1/3$, $\sigma_{\varphi}^2 = 1$. Our results are illustrated in Fig. ~\ref{fig:lr_experiment_indicator}.

	These experiments 
    demonstrate that the obtained bounds 
    capture well the decrease rate of the confidence intervals, only the outer approximation bound is a bit conservative.
	
	\section{Conclusions}\label{sec:conclusion}
	\vspace{-1.5mm}
    In this paper we have studied the sample complexity of the Sign-Perturbed Sums (SPS) finite-sample, distribution-free system identification method in the case of scalar linear regression problems. We have proved concentration bounds which show that the sizes of the SPS confidence intervals shrink at a geometric rate (around the true parameter). Similar results were proven for the outer approximation of the region. These results are directly relevant, e.g., for multi-armed bandits and various signal processing problems. Furthermore, the results and the applied techniques can serve as stepping stones to study the sample complexity of SPS in more complex multidimensional cases. Future research directions include extending the results to more general identification problems.
    \medskip
	
	\bibliography{ifacconf}             %

\begin{thebibliography}{19}
\providecommand{\natexlab}[1]{#1}
\providecommand{\url}[1]{\texttt{#1}}
\providecommand{\urlprefix}{URL }
\expandafter\ifx\csname urlstyle\endcsname\relax
  \providecommand{\doi}[1]{doi:\discretionary{}{}{}#1}\else
  \providecommand{\doi}{doi:\discretionary{}{}{}\begingroup
  \urlstyle{rm}\Url}\fi

\bibitem[{Baggio et~al.(2022)Baggio, Car{\`e}, Scampicchio, and
  Pillonetto}]{baggio2022bayesian}
Baggio, G., Car{\`e}, A., Scampicchio, A., and Pillonetto, G. (2022).
\newblock Bayesian frequentist bounds for machine learning and system
  identification.
\newblock \emph{Automatica}, 146.

\bibitem[{Bercu and Touati(2008)}]{bercu2008}
Bercu, B. and Touati, A. (2008).
\newblock Exponential inequalities for self-normalized martingales with
  applications.
\newblock \emph{The Annals of Applied Probability}, 18(5), 1848--1869.

\bibitem[{Campi and Weyer(2005)}]{Campi2005}
Campi, M.C. and Weyer, E. (2005).
\newblock Guaranteed non-asymptotic confidence regions in system
  identification.
\newblock \emph{Automatica}, 41, 1751--1764.

\bibitem[{Car\`e et~al.(2018)Car\`e, Cs\'aji, Campi, and Weyer}]{Algo2018}
Car\`e, A., Cs\'aji, B.{\relax{Cs}}., Campi, M., and Weyer, E. (2018).
\newblock Finite-sample system identification: An overview and a new
  correlation method.
\newblock \emph{IEEE Control Systems Letters}, 2(1), 61 -- 66.

\bibitem[{Carè et~al.(2021)Carè, Campi, Csáji, and Weyer}]{Care2021}
Carè, A., Campi, M., Csáji, B.{\relax Cs}., and Weyer, E. (2021).
\newblock Facing undermodelling in {Sign-Perturbed-Sums} system identification.
\newblock \emph{Systems and Control Letters}, 153, 104936.

\bibitem[{Carè(2022)}]{Care2022}
Carè, A. (2022).
\newblock A simple condition for the boundedness of {Sign-Perturbed-Sums (SPS)}
  confidence regions.
\newblock \emph{Automatica}, 139, 110150.

\bibitem[{Cs\'aji et~al.(2015)Cs\'aji, Campi, and Weyer}]{Csaji2015}
Cs\'aji, B.{\relax Cs}., Campi, M.C., and Weyer, E. (2015).
\newblock \relax{Sign-Perturbed Sums}: A new system identification approach for
  constructing exact non-asymptotic confidence regions in linear regression
  models.
\newblock \emph{IEEE Transactions on Signal Processing}, 63(1), 169--181.

\bibitem[{Cs\'aji and Weyer(2011)}]{Csaji2011}
Cs\'aji, B.{\relax Cs}. and Weyer, E. (2011).
\newblock System identification with binary observations by stochastic
  approximation and active learning.
\newblock In \emph{Proceedings of the 50th IEEE CDC and ECC, Orlando, Florida,
  December 12-15}.

\bibitem[{Cs\'aji and Weyer(2015)}]{Csaji2015cdc}
Cs\'aji, B.{\relax Cs}. and Weyer, E. (2015).
\newblock Closed-loop applicability of the \relax{Sign-Perturbed Sums} method.
\newblock In \emph{54th IEEE Conference on Decision and Control}, 1441--1446.

\bibitem[{Cs{\'a}ji and Horv{\'a}th(2022)}]{csaji2022nonparametric}
Cs{\'a}ji, B.{\relax Cs}. and Horv{\'a}th, B. (2022).
\newblock Nonparametric, nonasymptotic confidence bands with {Paley-Wiener}
  kernels for band-limited functions.
\newblock \emph{IEEE Control Systems Letters}, 6, 3355--3360.

\bibitem[{Cs{\'a}ji and Kis(2019)}]{csaji2019distribution}
Cs{\'a}ji, B.{\relax Cs}. and Kis, K.B. (2019).
\newblock Distribution-free uncertainty quantification for kernel methods by
  gradient perturbations.
\newblock \emph{Machine Learning}, 108(8), 1677--1699.

\bibitem[{Kolumb{\'a}n et~al.(2015)Kolumb{\'a}n, Vajk, and
  Schoukens}]{kolumban2015perturbed}
Kolumb{\'a}n, S., Vajk, I., and Schoukens, J. (2015).
\newblock Perturbed datasets methods for hypothesis testing and structure of
  corresponding confidence sets.
\newblock \emph{Automatica}, 51, 326--331.

\bibitem[{Lattimore and Szepesvári(2020)}]{lattimore_szepesvari_2020}
Lattimore, T. and Szepesvári, C. (2020).
\newblock \emph{Bandit Algorithms}.
\newblock Cambridge University Press.

\bibitem[{Ljung(1999)}]{Ljung1999}
Ljung, L. (1999).
\newblock \emph{System Identification: Theory for the User}.
\newblock Prentice Hall, Upper Saddle River, 2nd edition.

\bibitem[{S{\"o}derstr{\"o}m and Stoica(1989)}]{Soderstrom1989}
S{\"o}derstr{\"o}m, T. and Stoica, P. (1989).
\newblock \emph{System Identification}.
\newblock Prentice Hall International, Hertfordshire, UK.

\bibitem[{Tam{\'a}s and Cs{\'a}ji(2021)}]{tamas2021exact}
Tam{\'a}s, A. and Cs{\'a}ji, B.{\relax Cs}. (2021).
\newblock Exact distribution-free hypothesis tests for the regression function
  of binary classification via conditional kernel mean embeddings.
\newblock \emph{IEEE Control Systems Letters}, 6, 860--865.

\bibitem[{Volpe et~al.(2015)Volpe, Cs\'aji, Car{\`e}, Weyer, and
  Campi}]{volpe2015sign}
Volpe, V., Cs\'aji, B.{\relax Cs}., Car{\`e}, A., Weyer, E., and Campi, M.C.
  (2015).
\newblock {S}ign-{P}erturbed {S}ums ({SPS}) with instrumental variables for the
  identification of \relax{ARX} systems.
\newblock In \emph{54th IEEE Conference on Decision and Control (CDC), Osaka,
  Japan}, 2115--2120.

\bibitem[{Wainwright(2019)}]{wainwright_2019}
Wainwright, M.J. (2019).
\newblock \emph{High-Dimensional Statistics: A Non-Asymptotic Viewpoint}.
\newblock Cambridge Univ.\ Press.

\bibitem[{Weyer et~al.(2017)Weyer, Campi, and Cs\'aji}]{Weyer2017}
Weyer, E., Campi, M.C., and Cs\'aji, B.{\relax Cs}. (2017).
\newblock Asymptotic properties of {SPS} confidence regions.
\newblock \emph{Automatica}, 81, 287--294.

\end{thebibliography}

\end{document}